\documentclass[11pt,a4paper]{article}

\usepackage[T1]{fontenc}
\usepackage[utf8]{inputenc}
\usepackage[margin=2.5cm]{geometry}
\usepackage{amsmath,amssymb,amsthm}
\usepackage{graphicx}
\usepackage{booktabs,tabularx,array}
\usepackage{xcolor}
\usepackage{listings}
\lstdefinestyle{kcjson}{
  basicstyle=\ttfamily\scriptsize,
  breaklines=true,
  showstringspaces=false,
  frame=single,
  rulecolor=\color{lightgray-rule},
  backgroundcolor=\color{kc-bg},
  columns=fullflexible,
  keepspaces=true,
  xleftmargin=4pt, xrightmargin=4pt,
  aboveskip=1em, belowskip=0.5em
}
\usepackage{enumitem}
\usepackage{hyperref,url}
\usepackage{natbib}
\usepackage{authblk}
\usepackage{abstract}
\usepackage{titlesec}
\usepackage{fancyhdr}
\usepackage{caption,float}

\theoremstyle{definition}

\newtheorem{proposition}{Proposition}

\definecolor{mondegreen}{RGB}{26,107,74}
\definecolor{codeblue}{RGB}{30,80,140}
\definecolor{codegray}{RGB}{90,90,90}
\definecolor{lightgray-rule}{RGB}{200,200,200}
\definecolor{kc-bg}{RGB}{248,249,251}

\hypersetup{
  colorlinks=true, linkcolor=mondegreen,
  citecolor=mondegreen, urlcolor=codeblue,
  pdftitle={Knowledge Cards: Structured Knowledge for AI Systems},
  pdfauthor={Liliana Ferreira}
}

\titleformat{\section}{\large\bfseries}{}{0em}{\thesection\quad}
\titleformat{\subsection}{\normalsize\bfseries\itshape}{}{0em}{\thesubsection\quad}

\title{
  \vspace{-0.5cm}
  \LARGE\textbf{Knowledge Cards:\\
  Structured Knowledge for AI Systems}
}
\author[1]{Liliana Ferreira%
  \thanks{Corresponding author:
    \href{mailto:liliana.ferreira@mondegreen.ai}{liliana.ferreira@mondegreen.ai}}}
\affil[1]{Mondegreen.ai, Porto, Portugal}
\date{July 2026}

\begin{document}
\maketitle\thispagestyle{fancy}

\begin{abstract}
\noindent
AI systems whose outputs inform real decisions, and increasingly consequential ones, require something that current documentation practice does not provide: a structured, inspectable representation of the knowledge they need to ground, contextualize, and reason about those decisions, ideally reviewed and signed off by a domain expert. 
Established documentation artefacts already capture important aspects of an AI system. Model cards describe how a system behaves, data cards describe what it was trained on, and system cards describe the risks of a deployed system. None of them addresses the layer between inputs and outputs, more precisely, the concepts a system holds, the relationships it models, and the patterns of reasoning it applies. For pattern-recognition tasks this gap is tolerable.
For agentic AI, where systems act on their conclusions, it is the step that
most often separates a promising proof of concept from an operational solution
an organisation can rely on.

This paper introduces the Knowledge Card, a structured artefact that captures
validated knowledge about a single bounded concept in a form that experts can
review, organisations can audit, and AI systems can reason over. For one
concept, such as a specific failure mode, a compliance obligation, or a process
decision, a Knowledge Card records the entities and relationships involved,
the reasoning that connects them, the conditions under which that reasoning no longer holds, and the provenance of every claim, all grounded in a formal
domain ontology and signed off by a domain expert. Five properties define it:
ontology-grounded, provenance-linked, boundary-explicit, expert-validated, and
versioned. The manuscript develops the architecture, illustrates it with an agentic
AI case in the energy sector, and discusses three settings the same structure
supports: agent memory, industrial diagnostics, and regulatory compliance. Initial prototype cards have been built in the energy and pharmaceutical domains. The
schema is released as a public draft for community engagement.

\end{abstract}

\vspace{0.5em}\noindent\rule{\linewidth}{0.4pt}\vspace{0.5em}

\section{Introduction}
\label{sec:intro}

The documentation of AI systems has matured along several dimensions. Model
cards~\citep{mitchell2019} describe behavioural properties: what a system does,
how it performs across groups, what its known limitations are. Data
cards~\citep{pushkarna2022} describe input properties: provenance, collection
methodology, biases. System cards extend this to the risks of a deployed
system as a whole. These artefacts have become standard practice in research
and increasingly in industry.

A further layer has not been documented at all: the knowledge a system reasons
from, the concepts it holds, the relationships it models, the reasoning
patterns it applies, and the limits beyond which that reasoning no longer
holds. For pattern-recognition systems the omission is acceptable. However, this knowledge layer becomes
decisive for agentic AI, where a system does not merely classify or retrieve
but acts on its conclusions: dispatches a crew, approves a change, adjusts a
control. Here the company deploying the agent must be able to see and stand behind the knowledge it reasons from. This is the layer in which many AI initiatives stall. In fact, a proof of concept performs well in a demonstration but
cannot be moved into production, because the knowledge the system relies on
cannot be inspected, validated, or governed. More recently, industry analysts describe a
\textit{semantic layer} of this kind as a foundational requirement for enterprise
AI rather than an optimisation~\citep{gartner2026semantic}.

Consider an agent that helps a wind-farm operator act on a gearbox vibration
alert. The agent must decide whether the signal indicates genuine bearing
degradation, requesting an expensive maintenance dispatch, or a temporary fluctuation that can be safely watched. To make that call responsibly it needs
more than retrieved manuals: it needs to know which signals are decisive, how
they relate causally, and the conditions under which its own diagnosis does
not apply. 

Several recent approaches enrich what an AI system can draw on at run time.
Retrieval-augmented generation fetches relevant text and lets the model
interpret it. Graph-augmented retrieval~\citep{edge2024graphrag} structures a
corpus into a knowledge graph; documentation formats such as
DeepWiki~\citep{deepwiki2025} organise material into linked pages that agents
can browse. These approaches answer important questions: what relevant material exists, and how to surface it efficiently. But surfacing material is not the same as supplying knowledge. It is useful here to distinguish data, as raw records and signals, from information, which is data placed in context, and from knowledge, which is information that has been validated, related to what else is known, and made fit to reason and act upon. Retrieval and graph-augmented retrieval operate on data and information: they make the right content easier to find. A different question remains: whether the knowledge an agent needs, the knowledge that is validated, bounded, and able to be inspected and stood behind, is actually available to it. Retrieval makes content easier to find; it does not, on its own, make that content authoritative. 

This paper introduces the Knowledge Card, a structured artefact that captures
validated knowledge about a single bounded concept in a form that experts can
review, organisations can audit, and AI systems can reason over. For a given
concept, a Knowledge Card records the entities and relationships involved, the
reasoning that connects them, the conditions under which that reasoning no
longer holds, and the provenance of every claim. All of this is grounded in a
formal domain ontology, so each term carries a fixed meaning, and signed off
by a domain expert before the card is treated as authoritative. Five
properties, further introduced in Section~\ref{sec:kc}, make this precise:
ontology-grounded, provenance-linked, boundary-explicit, expert-validated, and
versioned.

A card is authored with a domain expert in the
loop. While often drafted by an automated pipeline, the knowledge card is designed to be corrected and
approved by the expert, and afterwards consumed by machines. An
LLM-based agent can read a card directly and reason over it in natural
language; because every term resolves to a formal ontology, the same card can
also be checked by a symbolic reasoner. A Knowledge Card thus complements the
use of LLMs: it gives an agent validated knowledge to reason from, and gives
an organisation a way to verify that the reasoning stayed within known bounds.

The contributions of this paper are:

\begin{itemize}[leftmargin=1.5em,itemsep=0.2em]
  \item The Knowledge Card is introduced as a structured artefact for
        validated knowledge, defined by five properties: ontology-grounded,
        provenance-linked, boundary-explicit, expert-validated, and versioned.
  \item The architecture is grounded in the separation between a domain
        ontology, the shared, formal vocabulary of a field, and the
        individual cards that instantiate it.
  \item The same structure applies to any bounded reasoning task: a
        failure mode, a compliance check, a process decision. A Knowledge Card
        is not tied to one domain or one kind of knowledge; it is a general
        container for a single, scoped, validated unit of reasoning. This
        generality is what lets a library of cards serve as a semantic layer
        across an organisation's AI systems rather than a one-off artefact.
  \item Three settings are discussed: agent memory, industrial diagnostics,
        and regulatory compliance. In each of these, Knowledge Cards meet needs that
        existing artefacts do not.
  \item The v0.1 schema is released as a public draft at
        \url{https://github.com/mondegreenai/KnowledgeCard}.
\end{itemize}

\noindent The remainder of the paper is organised as follows.
Section~\ref{sec:related} reviews related work: AI documentation artefacts,
retrieval and graph-based approaches, and the contemporary formats closest to
the Knowledge Card. Section~\ref{sec:kc} describes the Knowledge Card and its
five defining properties, using a worked energy-sector example as a guide.
Section~\ref{sec:schema} presents the schema and the reasoning
layer. Section~\ref{sec:applications} discusses three settings the structure
supports, and Section~\ref{sec:conclusion} concludes.

\section{Related Work}
\label{sec:related}

This section situates the Knowledge Card among three bodies of work. First,
the AI documentation artefacts, as the model, data, and system cards, that
initiated the practice of structured documentation but stop short of the
knowledge layer. Second, retrieval and graph-based approaches that improve how
relevant material reaches a model at run time. Third, the contemporary formats
closest in spirit to the Knowledge Card, which converge on the same gap from
different directions.

\subsection{AI Documentation Artefacts}

Model cards~\citep{mitchell2019} were introduced to make a model's intended
use, performance across groups, and known limitations explicit, so that a
system could be evaluated and adopted responsibly. Data
cards~\citep{pushkarna2022} did the same for datasets, documenting provenance
and collection methodology. IBM AI FactSheets~\citep{hind2019} and system
cards extended the practice to trustworthiness and deployment risk. Together
they established a valuable principle: an AI system should travel with
structured documentation of how it was built and how it behaves.

What these artefacts describe is the system and its data. The knowledge the
system reasons from, the specific concepts, causal relationships, and
inference patterns it applies to a given case, sits at a different level,
and no existing artefact class captures it. A model card may note a system's
limitations in prose, at the level of the system as a whole; it does not
represent, in machine-readable and formally grounded form, the reasoning the
system performs on a particular concept. The transparency obligations of the
EU AI Act~\citep{euaiact2024} and the FDA AI/ML framework~\citep{fda2021}
reach into exactly this layer, more concretely the conceptual foundations and limits of a
system's reasoning, which is why the gap matters in practice.

\subsection{Retrieval, Graph-Augmented Retrieval, and Context Graphs}

A large body of work improves how relevant data and information reach a model at run time. GraphRAG~\citep{edge2024graphrag} builds a knowledge graph from a
document corpus and reports gains in the comprehensiveness and diversity of answers to broad, sense-making queries, relative to conventional retrieval. LlamaIndex Property
Graph~\citep{llamaindex2024pg} and Neo4j GraphRAG~\citep{neo4j2024graphrag}
provide related graph-construction and retrieval tooling. These systems
organise a corpus so that an LLM can find and combine relevant passages more
effectively. Their graphs are built automatically and labelled with strings;
the meaning of a node or an edge is whatever the extracting model inferred,
and may not be fixed against a formal ontology~\citep{edge2024graphrag}.

Two contemporary systems sit closer to the Knowledge Card and are worth a
direct comparison, because both engage the same underlying gap: the absence of
a shared standard for feeding validated knowledge into AI and agentic systems.

The TrustGraph context graph~\citep{davis2025contextgraph} extends a knowledge
graph with OWL-grounded entity types, PROV-O provenance to source documents,
and temporal awareness. The Open Knowledge Format
(OKF)~\citep{mcveety2026okf}, released by Google Cloud, takes the opposite
design stance: knowledge is a directory of linked markdown files with YAML
frontmatter, with a single mandatory field and the content model left to the
producer. Table~\ref{tab:compare} summarises how these and the retrieval
systems relate to a Knowledge Card. The comparison identifies what an agentic system most needs from a knowledge source: a unit of knowledge that is formally grounded, supports reasoning and verification, declares its own scope, and carries an explicit record of expert validation.

\begin{table}[h]
\centering
\caption{How related approaches compare on the properties an agentic system
needs from a knowledge source. ``Formal grounding'' means terms are fixed
against an ontology rather than interpreted per consumer; ``supports
reasoning'' means a symbolic reasoner can verify conclusions, not only an LLM;
``scope declared'' means the artefact states where it ceases to apply.}
\label{tab:compare}
\renewcommand{\arraystretch}{1.3}
\small
\begin{tabular}{@{}lcccc@{}}
\toprule
 & \textbf{Formal} & \textbf{Supports} & \textbf{Scope} & \textbf{Expert} \\
 & \textbf{grounding} & \textbf{reasoning} & \textbf{declared} & \textbf{validation} \\
\midrule
GraphRAG / property graphs & no & partial & no & no \\
OKF (markdown bundles)     & no & via LLM only & no & no \\
TrustGraph context graph   & yes & partial & no & no \\
\textbf{Knowledge Card}    & \textbf{yes} & \textbf{yes} & \textbf{yes} & \textbf{yes} \\
\bottomrule
\end{tabular}
\end{table}

A Knowledge Card contributes to the creation of a shared structure, so that the same concept means the same thing
across every system that consumes it.

\section{The Knowledge Card}
\label{sec:kc}

A Knowledge Card captures validated knowledge about a bounded concept.
It is a named individual in a domain knowledge base: it draws its vocabulary
from a formal domain ontology. An ontology represents the shared definitions of what a failure
mode is, what a process parameter is, and what it means for one to affect another. The Knowledge Card instantiates that vocabulary for one specific concept. Because the
vocabulary is formal, every claim on a card inherits the inferential machinery
of the ontology: terms have fixed meanings, relationships have declared
domains and ranges, and conclusions can be checked for consistency. This is
what separates a Knowledge Card from a document or an automatically extracted
graph, where a label means only what a reader or a model takes it to mean.

A Knowledge Card $\mathit{KC}$ for concept $C$ in system $S$ is the tuple:
\begin{equation}
  \mathit{KC}(C, S) =
    \langle \mathcal{M},\, \mathcal{C},\, \mathcal{E},\, \mathcal{R},\,
            \mathcal{P},\, \mathcal{B},\, \mathcal{L} \rangle
\end{equation}
where $\mathcal{M}$ is metadata (provenance, lifecycle, validation record),
$\mathcal{C}$ is the concept anchor (an ontology class), $\mathcal{E}$ is the
entity register (the individuals involved, with typed roles), $\mathcal{R}$ is
the relationship topology (named relationships with confidence scores),
$\mathcal{P}$ is the reasoning pattern set, $\mathcal{B}$ is the boundary
condition set, and $\mathcal{L}$ is the link to the consuming system.

\subsection{A Worked Example: A Wind-Farm Maintenance Agent}
\label{sec:example}

Consider the agent from the introduction, supporting a wind-farm operator. A
gearbox vibration alert has fired on a turbine. The agent's task is to decide
whether the alert indicates outer-race spalling on the high-speed-shaft
bearing, a degradation mode that warrants dispatching a maintenance crew, or a benign transient. A Knowledge Card for this concept gives the agent
the validated structure it needs to reason responsibly.

The failure mode
\texttt{energy:GearboxBearing\_OuterRaceSpalling} is the card's concept anchor. Its entity register lists
what matters and how: the high-speed-shaft bearing (the subject), the bearing
outer race (the component that degrades), and the observable signals, namely the
vibration amplitude at the bearing's characteristic fault frequency, the oil
temperature trend, and the ferrous particle count in the lubricant. Each signal is
tagged with how directly it can be observed and how much it contributes to the
diagnosis. Its relationship topology records the causal structure: outer-race
degradation generates the characteristic vibration signature and raises the
particle count. Its reasoning patterns range from a plain-language description
for a human reviewer to a formal rule a symbolic reasoner can check. And,
critically, its boundary conditions state where the diagnosis does not apply. For instance, if the bearing was re-greased within the previous 48 hours,
the vibration signature is unreliable, and the card declares itself
inapplicable, so the agent escalates to a human rather than recommending a
costly dispatch on false evidence.

This single card is enough to take the agent from a vague awareness that manuals mention
bearing faults to a precise judgement: this signal pattern, in this context, indicates
outer-race spalling with calibrated confidence, unless one of these stated
conditions holds, in which case the agent should not decide alone.

The rest of this
section uses this example to make the five properties concrete.

\subsection{Five Defining Properties}

Five properties distinguish a Knowledge Card from a retrieved document or an
automatically extracted graph. 

\textit{Ontology-grounded.} Every entity and relationship in the card is
anchored to a formal ontology term, not a free-text label. In the example,
``outer-race spalling'' is a defined class with known relationships to other
failure modes, not a string that a model might confuse with a different fault.
This way, the same concept means the same thing across every agent, report,
and system in the organisation, the property that lets a collection of
cards act as a shared semantic layer rather than a set of disconnected notes.

\textit{Provenance-linked.} Every claim traces to its source, the standard,
the maintenance record, the engineer who validated it, and to the version
of the ontology under which it was checked. Therefore, when an agent
recommends a dispatch, the organisation can see exactly which evidence and
whose judgement the recommendation rests on, which is what makes the output
auditable after the fact.

\textit{Boundary-explicit.} The card states the conditions under which its
reasoning ceases to hold, as first-class content. In the example, the
re-greasing condition is part of the card, not an afterthought. By ensuring this, the
agent knows the limits of its own competence and escalates rather than acting
on a diagnosis that does not apply. This is the property that most directly
prevents confident, costly mistakes.

\textit{Expert-validated.} A card reaches its authoritative state only after a
qualified domain expert reviews and signs it off, recorded in its metadata. A
draft generated by an automated pipeline is marked as such until that review
happens. The knowledge an agent reasons from is warranted by a
named human expert, not merely statistically plausible, which is what an
organisation needs in order to take responsibility for the agent's actions.

\textit{Versioned.} As evidence accumulates and the reasoning is refined, new
versions supersede old ones, and earlier decisions remain traceable to the
version that governed them. With this, the knowledge improves over time without
losing the audit trail, so a decision made last year can still be explained in
terms of what was known then.

\section{Schema and Reasoning}
\label{sec:schema}

The Knowledge Card schema (JSON Schema 2020-12 with a JSON-LD context) is
available at \url{https://github.com/mondegreenai/KnowledgeCard}.
Figure~\ref{fig:kc1} shows an abbreviated serialisation of the maintenance card
from Section~\ref{sec:example}, which the rest of this section refers to.

\begin{figure}
\centering
\begin{lstlisting}[style=kcjson]
{
  "@type": "kc:KnowledgeCard",
  "kc:concept": "energy:GearboxBearing_OuterRaceSpalling",
  "kc:meta": {
    "kc:lifecycle": "validated",
    "kc:version": "1.0",
    "kc:validatedBy": "reliability-engineer-anon",
    "kc:ontologyVersion": "energydiag-0.2"
  },
  "kc:entities": [
    { "@id": "energy:HSSBearing",        "kc:entityRole": "primary-subject",
      "kc:observability": "inferred",      "kc:criticalityWeight": 1.00 },
    { "@id": "energy:BPFOAmplitude",     "kc:entityRole": "observable-signal",
      "kc:observability": "sensor-derived","kc:criticalityWeight": 0.94 },
    { "@id": "energy:FerrousParticleCount","kc:entityRole": "observable-signal",
      "kc:observability": "sensor-derived","kc:criticalityWeight": 0.85 }
  ],
  "kc:relationships": [
    { "kc:from": "energy:BearingOuterRace", "kc:property": "generates",
      "kc:to": "energy:BPFOAmplitude",      "kc:confidence": 0.93 }
  ],
  "kc:reasoning": {
    "kc:level3": {
      "kc:ruleBody": "HSSBearing(?b) ^ hasBPFOAmplitude(?b,?a) ^ greaterThan(?a,2.0)
                      ^ hasFerrousTrend(?b,'rising') ^ hasOperatingMode(?b,'rated')
                      -> hasFailureMode(?b, energy:GearboxBearing_OuterRaceSpalling)",
      "kc:calibratedConfidence": 0.89
    }
  },
  "kc:boundaries": [
    { "kc:severity": "card-invalid",
      "kc:condition": "bearing re-greased within previous 48 hours" },
    { "kc:severity": "card-invalid",
      "kc:condition": "turbine operating outside rated wind-speed band" }
  ]
}
\end{lstlisting}
\caption{Abbreviated Knowledge Card for the wind-farm maintenance example.}
\label{fig:kc1}
\end{figure}

\subsection{Schema Fields}

The fields follow the tuple of Section~\ref{sec:kc}. \texttt{kc:meta} carries
provenance, lifecycle state, and the validation record. \texttt{kc:concept}
anchors the card to an ontology class. \texttt{kc:entities} registers each
individual with its role (subject, causal agent, causal target, observable
signal, contextual factor), how observable it is, and a criticality weight in
$[0,1]$. \texttt{kc:relationships} records named relationships with a
confidence in $[0,1]$. \texttt{kc:reasoning} carries the layered reasoning of
Section~\ref{subsec:reasoning}; the narrative and rule layers are expected, and
the probabilistic layer is optional, present only when calibration data supports it. \texttt{kc:boundaries} lists scope limits with a
severity rating, and \texttt{kc:systemLink} binds the card to the consuming
system. For brevity, \texttt{kc:systemLink} is omitted from the abbreviated card in Figure~\ref{fig:kc1}.

\subsection{Reasoning}
\label{subsec:reasoning}

A Knowledge Card represents its reasoning in layers, from a form any reader can
follow to a form a machine can check. The layers describe the same inference at
increasing levels of formality, so a card stays readable by a domain expert
while remaining executable by a system.

The layers are a ladder, not a checklist. A card carries as many of them as its
knowledge supports, and no more. An expert can almost always state the inference
in plain language and, in most cases, as a formal rule; a card with those two
layers is complete, valid, and useful on its own. The layers above and below are
added when the knowledge is there to support them.

The first layer is a plain-language description of how the conclusion follows
from the evidence. The second makes this a structured pattern: the signals,
their thresholds, and the conclusion they support, expressed so that a system
can match them against incoming data. The third states the inference as a
formal rule whose terms are ontology classes, so that a symbolic reasoner can
apply it and verify that a conclusion is warranted rather than merely
plausible. Figure~\ref{fig:reasoning} shows these layers for the wind-farm
card.

\begin{figure}[ht]
\centering
\begin{lstlisting}[style=kcjson]
"kc:reasoning": {

  "kc:level1_narrative":
    "A rising vibration amplitude at the bearing's outer-race fault
     frequency, together with an increasing ferrous particle count in
     the lubricant, indicates outer-race spalling on the high-speed-shaft
     bearing. Oil temperature trend corroborates. The diagnosis holds only
     while the turbine operates within its rated wind-speed band.",

  "kc:level2_pattern": {
    "kc:requiredSignals": [
      { "@id": "energy:BPFOAmplitude",       "kc:threshold": "> 2.0 g",  "kc:trend": "rising" },
      { "@id": "energy:FerrousParticleCount", "kc:threshold": "> 15 ppm", "kc:trend": "rising" }
    ],
    "kc:corroboratingSignal": { "@id": "energy:OilTempTrend", "kc:trend": "rising" },
    "kc:conclusion": "energy:GearboxBearing_OuterRaceSpalling"
  },

  "kc:level3_rule": {
    "kc:ruleLanguage": "SWRL",
    "kc:ruleBody":
      "HSSBearing(?b) ^ hasBPFOAmplitude(?b,?a) ^ greaterThan(?a,2.0)
       ^ hasFerrousTrend(?b,'rising') ^ hasOperatingMode(?b,'rated')
       -> hasFailureMode(?b, energy:GearboxBearing_OuterRaceSpalling)",
    "kc:calibratedConfidence": 0.89
  }
}
\end{lstlisting}
\caption{The layered \texttt{kc:reasoning} block for the wind-farm card. The same diagnosis is expressed as a narrative, a structured pattern, and a formal rule.}
\label{fig:reasoning}
\end{figure}

A fourth layer adds probability, when the data exist to support it. Where
calibration data is available, the probabilistic layer captures what a
deterministic rule cannot, that the same signal carries different weight in
different contexts, and that conflicting signals should shift a degree of belief
rather than force a verdict. A card's causal structure maps naturally onto a Bayesian
network~\citep{pearl1988,jensen2001bn}: the relationships form the directed
graph, the confidence scores are its parameters, the observable signals are
evidence, and the conclusion is the variable being queried. Bayesian networks
have been the standard probabilistic tool in industrial diagnostics for three
decades~\citep{langseth2007bn}; a card gives them a portable container with
provenance, validation status, and explicit boundaries. Where the causal
structure is only partly known, or where rules must be softened to weigh
conflicting evidence, the same information can be expressed as a Markov Logic
Network~\citep{richardson2006}, with a formal rule of calibrated precision
$c \in (0,1)$ mapped to a weighted formula of weight $\log c / (1 - c)$; the
Bayesian network is recovered as the special case of a fully specified,
acyclic model.

The layers are also selective rather than uniform. A single card need not reason
one way throughout: it can apply a formal rule where the logic is crisp, fall
back to the probabilistic layer where the evidence is uncertain, and defer to a
human where neither holds. Complex problems are rarely settled by one method, and
rarely by one card. A hard diagnosis may be covered by several cards, each bounded
to the part of the problem it can answer, with their boundary conditions routing a
case from one to the next or to a person. The card does not pretend to a single
mechanism that fits every case; it carries the methods that fit, and it knows
which one applies.

The reasoning layers and the card's boundaries work together. A boundary
condition does not lower the probability of a conclusion, it removes the
card's authority to draw one. Figure~\ref{fig:boundaries} shows the boundary
block for the wind-farm card.

\begin{figure}[ht]
\centering
\begin{lstlisting}[style=kcjson]
"kc:boundaries": [
  { "kc:severity": "card-invalid",
    "kc:condition": "bearing re-greased within previous 48 hours",
    "kc:rationale": "re-greasing transient mimics the spalling signature" },

  { "kc:severity": "card-invalid",
    "kc:condition": "turbine operating outside rated wind-speed band" },

  { "kc:severity": "requires-human-review",
    "kc:condition": "concurrent drivetrain alert present" }
]
\end{lstlisting}
\caption{The \texttt{kc:boundaries} block. A \texttt{card-invalid} condition withdraws the card's authority to conclude; a \texttt{requires-human-review} condition defers to a person.}
\label{fig:boundaries}
\end{figure}

This is made precise by giving a boundary of severity \texttt{card-invalid} a
weight large enough to override any combination of positive evidence.

\begin{proposition}[A boundary overrides the conclusion]
\label{prop:defeat}
If a \texttt{card-invalid} condition holds for a case, the probability of any
positive conclusion is zero, whatever the supporting evidence.
\end{proposition}

\noindent In the wind-farm example, if the bearing was re-greased within the
previous 48 hours, the re-greasing condition fires and the card withdraws its
diagnosis: however strong the vibration signal, the agent cannot conclude a
fault on this card's authority, and escalates to a human instead. A
\texttt{card-invalid} condition does not make the conclusion unlikely; it makes
the card inapplicable, which is a stronger and more useful guarantee, since the
system fails safe rather than guessing. The guarantee does not depend on the
probabilistic layer: for a card that reasons deterministically, a card-invalid
condition simply blocks the rule from firing, with the same effect.

\subsection{Portability}

The schema specifies what a valid card contains and what its contents license;
it leaves to the consuming system how those inferences are computed. This keeps a card portable. The same
card can be read directly by an LLM-based agent and reasoned over in natural
language, and also checked by a symbolic reasoner; the symbolic grounding is
what lets the agent's natural-language reasoning be verified against the card's
content rather than taken on trust. The LLM supplies flexible reasoning, the
symbolic layer supplies the guarantee.

\section{Applications}
\label{sec:applications}

The same structure serves any setting that needs validated, inspectable
knowledge a system can reason over. Three settings show its range.

\subsection{Agent Memory}

As agentic AI moves from pilots to production, one failure mode recurs: the organisation cannot say what its agent knows, or account for it. 
Much of the current effort goes into context, through larger windows, better retrieval, more sophisticated prompt assembly. Yet more text is not the same as knowledge an agent can reason over, and simply lengthening the context can even degrade performance~\citep{liu2024lost}.
An agent that acts needs knowledge it can be held to, validated, bounded, attributable knowledge, and not just text it retrieved. This is the gap in agent memory. The literature~\citep{bian2025,pan2024unifying} maps the working context, the retrieved documents, and the model's parameters; it does not yet name the part an organisation can stand behind. A library of Knowledge Cards names it: the grounded, expert-validated, bounded portion of what an agent knows, the part that can be cited and audited when a decision is challenged. For the sake of clarity, we do not propose that the agent writes its own knowledge. A card is authored with a domain expert in the loop; the agent reads from this library and reasons over it, but does not silently revise it.

\subsection{Industrial Diagnostics}

Return to the wind-farm agent. When a vibration alert fires, the agent reads the gearbox card and works down its reasoning layers: the formal rule confirms that the signal pattern matches outer-race spalling, the probabilistic layer yields a confidence of 0.89 given a rising particle count alongside the vibration, and the narrative layer supplies the rationale an engineer can act on. Before recommending a maintenance dispatch, the agent evaluates the boundary block. If the bearing was re-greased within the previous 48 hours, the card withdraws its diagnosis, and the agent defers the costly dispatch and flags the turbine for human review rather than acting on an unreliable signal. A single card thus serves three readers in one decision: a rationale for the engineer, an executable rule for the system, and an auditable record for the organisation, and includes the basis on which the agent declined to act.

\subsection{Regulatory Compliance}

The same structure handles compliance, where the concept is a rule rather than a fault. Take an obligation from pharmaceutical manufacturing, governed by Good Manufacturing Practice (GMP), the regulated quality framework that determines whether a medicine can be released to patients. Under its change-control requirement, a proposed change to a validated production process may proceed only if its risk assessment, approval signatures, and revalidation evidence are all present and current. A Knowledge Card for that obligation anchors to the requirement, registers those evidence types as its entities, and encodes the rule as its reasoning, so that an agent can determine whether a specific change request actually clears the bar and cite exactly which condition is missing when it does not. Its boundaries mark where the card stops deciding. A change of a class the card was not validated for is escalated, not waved through. Because diagnostics and compliance share one structure, an organisation can build a single semantic layer that serves both its operational and its compliance agents, instead of maintaining separate, incompatible knowledge bases. The same validated knowledge is reused wherever a decision needs it.

\section{Conclusion}
\label{sec:conclusion}

The documentation of AI systems has produced well-adopted artefacts for the
behavioural and data layers. The knowledge a system reasons from, more concretely the
concepts it holds, the relationships it models, the reasoning it applies, and
the limits of that reasoning, has not had a structured form. As AI moves
from pattern recognition to action, and as agents are asked to make decisions
an organisation must stand behind, this is the layer that increasingly decides
whether a system reaches production at all.

A Knowledge Card gives that layer a form. It captures validated knowledge
about a single bounded concept, grounded in a formal ontology, signed off
by an expert, with explicit boundaries and full provenance. Its five
properties: ontology-grounded, provenance-linked, boundary-explicit,
expert-validated, versioned, make it a unit of knowledge that an agent can
reason over and a human can audit. Because the structure is shared, a library
of cards becomes more than a collection of artefacts: it is a semantic layer
that lets every system in an organisation reason from the same validated
knowledge, the foundation that agentic AI is now widely held to
require~\citep{gartner2026semantic}.

The contribution is a standard, not a system: a specification of what
validated knowledge should look like as an artefact, independent of how it is
stored, retrieved, or served. We offer the Knowledge Card as a candidate standard for the knowledge layer that agentic AI requires, the layer that turns retrieved text into knowledge a system can be trusted to act on. The schema is released as a public draft, and
contributions are invited at
\url{https://github.com/mondegreenai/KnowledgeCard}. The AI systems an
organisation will be able to trust with consequential decisions are those that
can make their knowledge explicit, their reasoning checkable, and their limits
known. 

\bibliographystyle{plainnat}
\bibliography{references-v15}

\end{document}